\documentclass{article}

\usepackage{arxiv}

\usepackage[utf8]{inputenc} 
\usepackage[T1]{fontenc}    
\usepackage{hyperref}       
\usepackage{url}            
\usepackage{booktabs}       
\usepackage{amsfonts}       
\usepackage{nicefrac}       
\usepackage{microtype}      
\usepackage{lipsum}
\usepackage{amsmath}
\usepackage{subcaption}
\usepackage{graphicx}
\usepackage{authblk}

\title{A deep neural network tool for automatic segmentation of human body parts in natural scenes}

\author[1\thanks{Corresponding author's  e-mail address: patrick.mcclure@nih.gov}]{Patrick McClure}
\author[2]{Gabrielle Reimann}
\author[2]{Michal Ramot}
\author[1]{Francisco Pereira}
\affil[1]{Machine Learning Team, National Institute of Mental Health}
\affil[2]{Section on Cognitive Neuropsychology, National Institute of Mental Health}

\begin{document}
\maketitle

\begin{abstract}

This short article describes a deep neural network trained to perform automatic segmentation of human body parts in natural scenes. More specifically, we trained a Bayesian SegNet with concrete dropout on the Pascal-Parts dataset to predict whether each pixel in a given frame was part of a person’s hair, head, ear, eyebrows, legs, arms, mouth, neck, nose, or torso.
\end{abstract}

\section{Overview}

Our deep neural network (DNN) tool was built to segment human body parts (the hair, the head, the left and right arms, the left and right eyes, the left and right eyebrows, the left and right legs, the left and right arms, the mouth, the neck, and the torso) from images of natural scenes. The goal in building this tool was to enable fast, automatic segmentation of human body parts from video frames, for enabling the analysis of human eye tracking data while watching those videos. The code is available at \url{https://github.com/nih-fmrif/MLT_Body_Part_Segmentation}.

\section{Data}

We trained and tested the DNN using the training and test sets from the Pascal-Parts \cite{everingham2010pascal,chen2014detect}, respectively. Only the images containing the "person" object were used, and the part labels were remapped so combine left and right labels for the same body part (see Table \ref{tab:parts}). During training, the dataset was augmented by random horizontal flipping, contrast, saturation, brightness, and hue of training sets images, as implemented in the torchvision transforms package \footnote[1]{\url{https://pytorch.org/docs/stable/torchvision/transforms.html}}.

\begin{table}[h]
 \caption{Segmented parts names and labels}
  \centering
  \tiny
  \begin{tabular}{lll}
    \toprule
    Pascal Part Name & New Part Name & New Part Label  \\
    \midrule
    Hair & Hair  &  1  \\
    Head     & Head & 2  \\
    Left Ear     & Ear       & 3 \\
    Left Eye     & Eye       & 4 \\
    Left Eyebrow     & Eyebrow       & 5 \\
    Left Foot     & Leg       & 6 \\
    Left Hand     & Arm       & 7 \\
    Left Lower Arm     & Arm       & 7 \\
    Left Lower Leg     & Leg       & 6 \\
    Left Upper Arm     & Arm       & 7 \\
    Left Upper Leg     & Leg       & 6 \\
    Right Ear     & Ear       & 3 \\
    Right Eye     & Eye       & 4 \\
    Right Eyebrow     & Eyebrow       & 5 \\
    Right Foot     & Foot       & 6 \\
    Right Hand     & Arm       & 7 \\
    Right Lower Arm     & Arm       & 7 \\
    Right Lower Leg     & Leg       & 6 \\
    Right Upper Arm     & Arm       & 7 \\
    Right Upper Leg     & Leg       & 6 \\
    Mouth     & Mouth & 8  \\
    Neck     & Neck & 9  \\
    Nose     & Nose & 10  \\
    Torso & Torso  &  11  \\
    Non-person objects & Background & 0 \\
    Background & Background & 0 \\
    \bottomrule
  \end{tabular}
  \label{tab:parts}
\end{table}

\section{Deep Neural Network}

We used PyTorch \cite{paszke2019pytorch} and the Pascal-Parts training dataset to train a Bayesian SegNet \cite{kendall2015bayesian}, with concrete dropout \cite{gal2017concrete} at the center and output layers, to perform automatic segmentation of human body parts from natural scenes. The detailed DNN architecture is shown in Table \ref{tab:arch}.

\begin{table}[h]
 \caption{Bayesian SegNet Architecture}
  \centering
  \tiny
  \begin{tabular}{lllll}
    \toprule
    Layer & Kernel Size & \#Features & Activation Function & Normalization/Dropout  \\
    \midrule
    Convolution 1 & 3x3 & 64  & ReLU & Batch Normalization  \\
    Convolution 2 & 3x3 & 64 & ReLU & Batch Normalization  \\
    Pool 1 & 2x2 & - & Max & - \\
    Convolution 3 & 3x3 & 128 & ReLU & Batch Normalization \\
    Convolution 4 & 3x3 & 128 & ReLU & Batch Normalization \\
    Pool 2 & 2x2 & - & Max & -  \\
    Convolution 5 & 3x3 & 256 & ReLU & Batch Normalization \\
    Convolution 6 & 3x3 & 256 & ReLU & Batch Normalization \\
    Convolution 7 & 3x3 & 256 & ReLU & Batch Normalization \\
    Pool 3 & 2x2 & - & Max & -  \\
    Convolution 8 & 3x3 & 512 & ReLU & Batch Normalization \\
    Convolution 9 & 3x3 & 512 & ReLU & Batch Normalization \\
    Convolution 10 & 3x3 & 512 & ReLU & Batch Normalization \\
    Pool 4 & 2x2 & - & Max & -  \\
    Convolution 11 & 3x3 & 512 & ReLU & Batch Normalization \\
    Convolution 12 & 3x3 & 512 & ReLU & Batch Normalization \\
    Convolution 13 & 3x3 & 512 & ReLU & Batch Normalization, Concrete Dropout \\
    Pool 5 & 2x2 & - & Max & -  \\
    Unpool 1 & 2x2 & - & Max & -  \\
    Convolution 14 & 3x3 & 512 & ReLU & Batch Normalization \\
    Convolution 15 & 3x3 & 512 & ReLU & Batch Normalization \\
    Convolution 16 & 3x3 & 512 & ReLU & Batch Normalization \\
    Unpool 2 & 2x2 & - & Max & -  \\
    Convolution 17 & 3x3 & 512 & ReLU & Batch Normalization \\
    Convolution 18 & 3x3 & 512 & ReLU & Batch Normalization \\
    Convolution 19 & 3x3 & 512 & ReLU & Batch Normalization \\
    Unpool 3 & 2x2 & - & Max & -  \\
    Convolution 20 & 3x3 & 256 & ReLU & Batch Normalization \\
    Convolution 21 & 3x3 & 256 & ReLU & Batch Normalization \\
    Convolution 22 & 3x3 & 256 & ReLU & Batch Normalization \\
    Unpool 4 & 2x2 & - & Max & -  \\
    Convolution 23 & 3x3 & 128 & ReLU & Batch Normalization \\
    Convolution 24 & 3x3 & 128 & ReLU & Batch Normalization \\
    Unpool 5 & 2x2 & - & Max & -  \\
    Convolution 25 & 3x3 & 64  & ReLU & Batch Normalization, Concrete Dropout  \\
    Convolution 26 & 3x3 & 64 & Linear &  -  \\
    
    \bottomrule
  \end{tabular}
  \label{tab:arch}
\end{table}

\pagebreak

\section{Results}

\begin{figure}[h!]
  \centering
  \begin{subfigure}[b]{0.49\textwidth}
  \includegraphics[width=\textwidth]{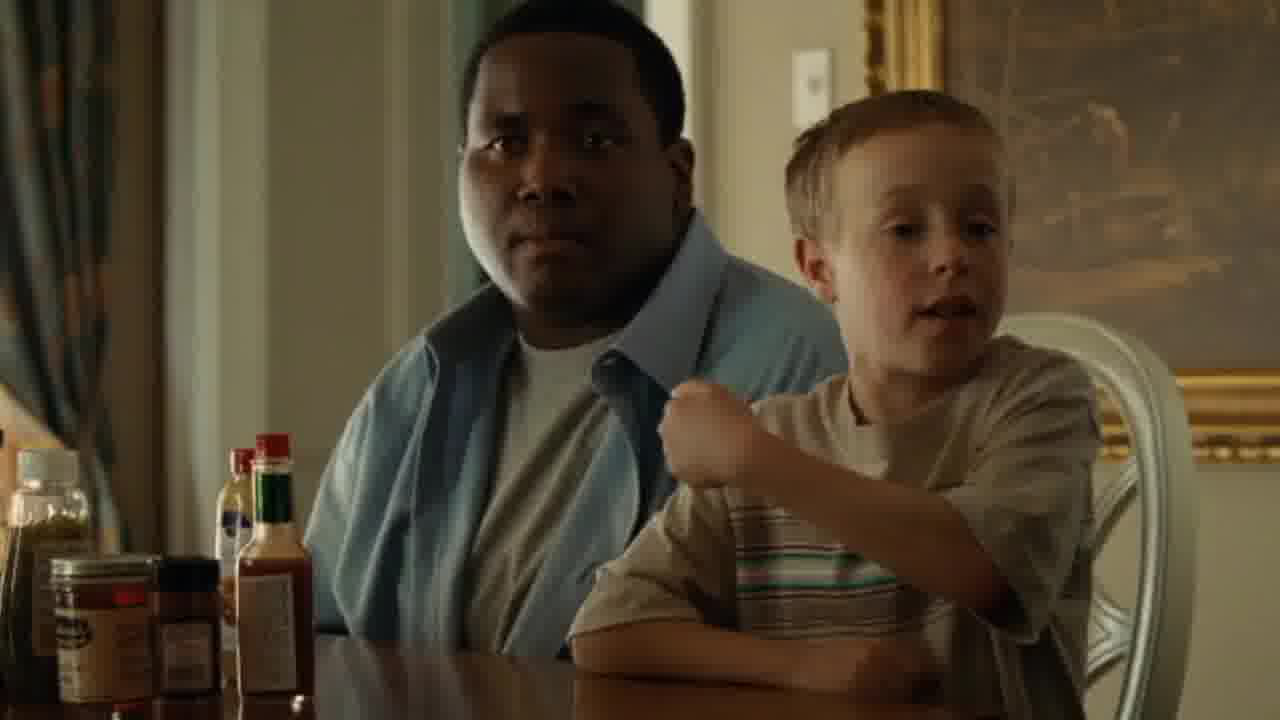}
  \end{subfigure}
  \begin{subfigure}[b]{0.49\textwidth}
  \includegraphics[width=\textwidth]{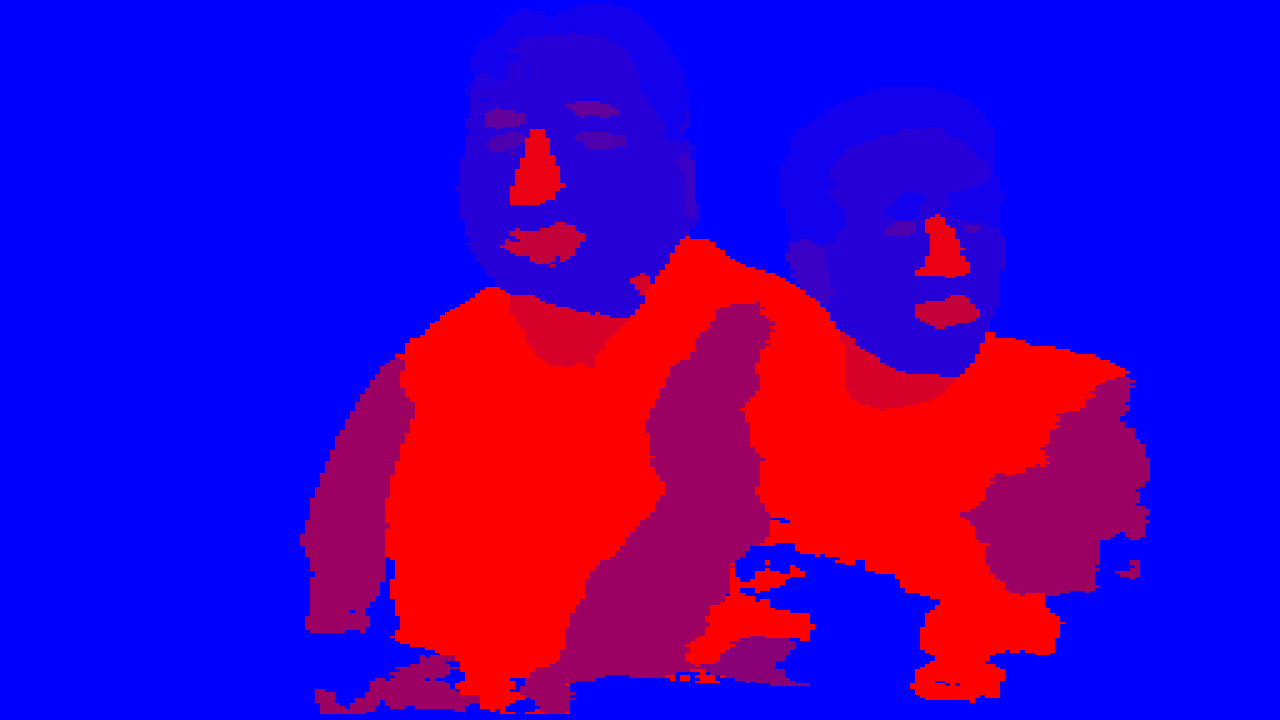}
  \end{subfigure}
  \begin{subfigure}[b]{\textwidth}
  \caption{}
  \end{subfigure}

  \begin{subfigure}[b]{0.49\textwidth}
  \includegraphics[width=\textwidth]{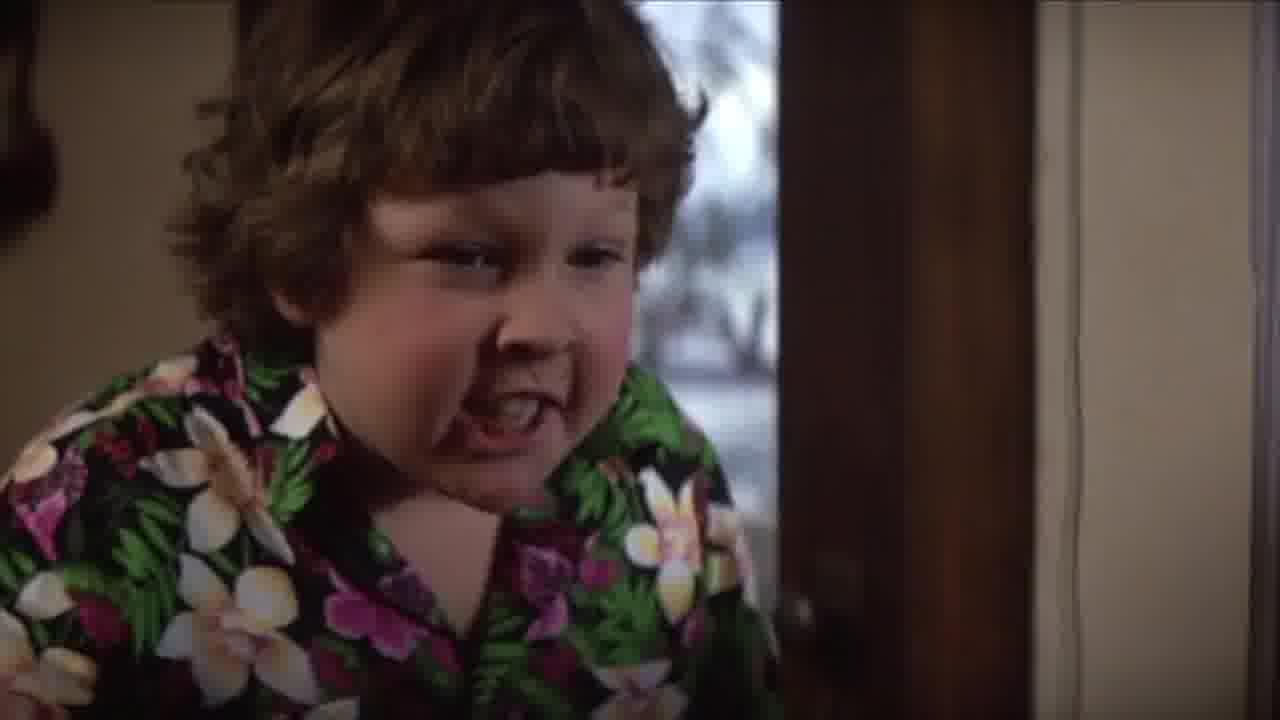}
  \end{subfigure}
  \begin{subfigure}[b]{0.49\textwidth}
  \includegraphics[width=\textwidth]{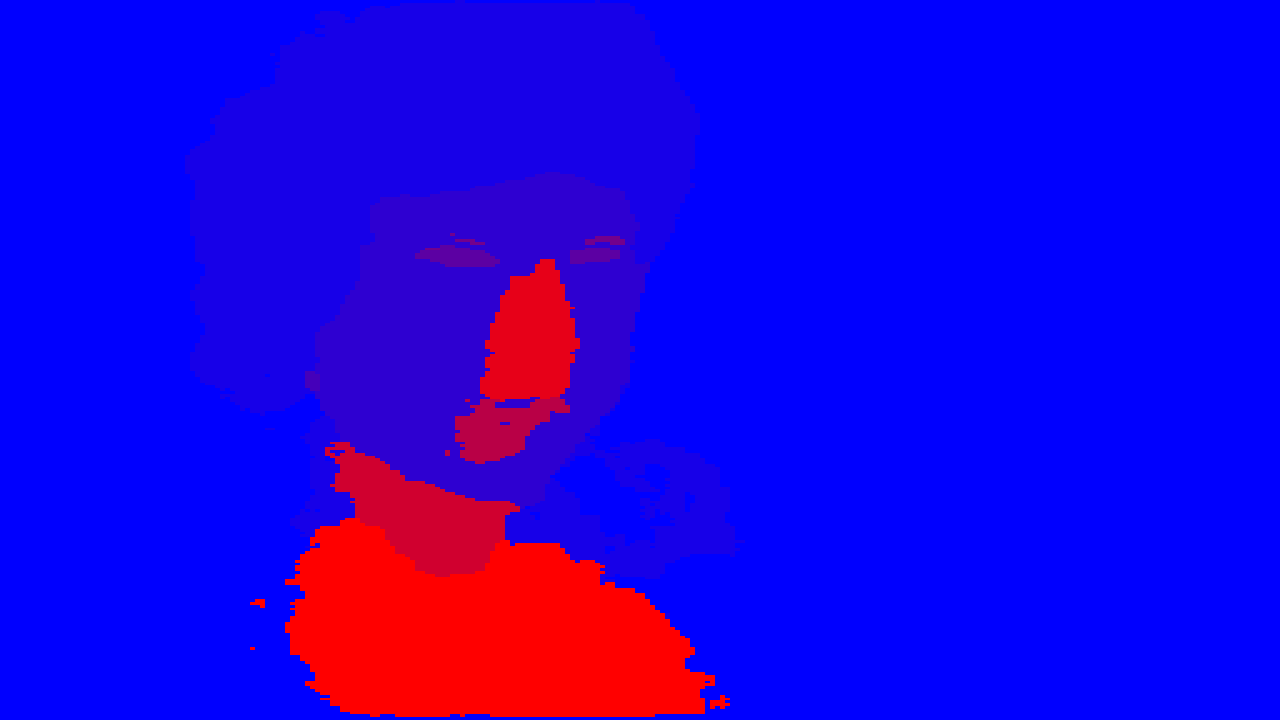}
  \end{subfigure}
    \begin{subfigure}[b]{\textwidth}
  \caption{}
  \end{subfigure}

  \begin{subfigure}[b]{0.49\textwidth}
  \includegraphics[width=\textwidth]{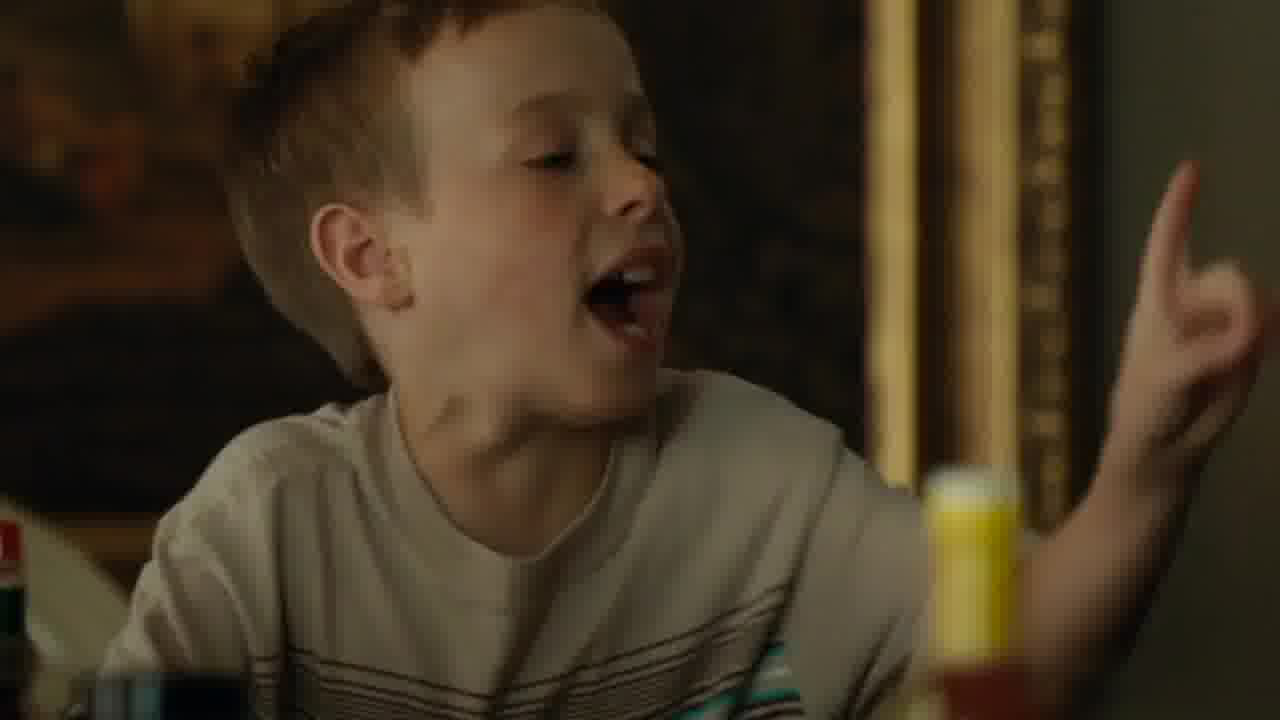}
  \end{subfigure}
  \begin{subfigure}[b]{0.49\textwidth}
  \includegraphics[width=\textwidth]{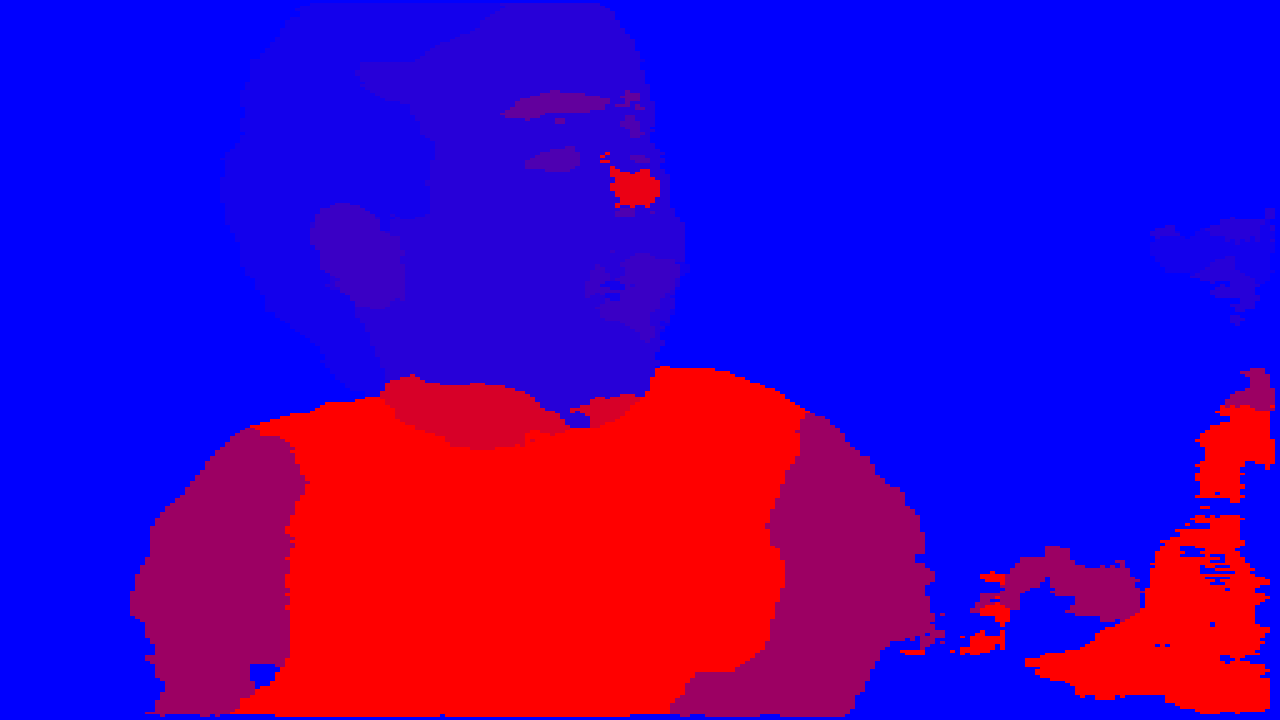}
  \end{subfigure}
    \begin{subfigure}[b]{\textwidth}
  \caption{}
  \end{subfigure}

  \begin{subfigure}[b]{0.49\textwidth}
  \includegraphics[width=\textwidth]{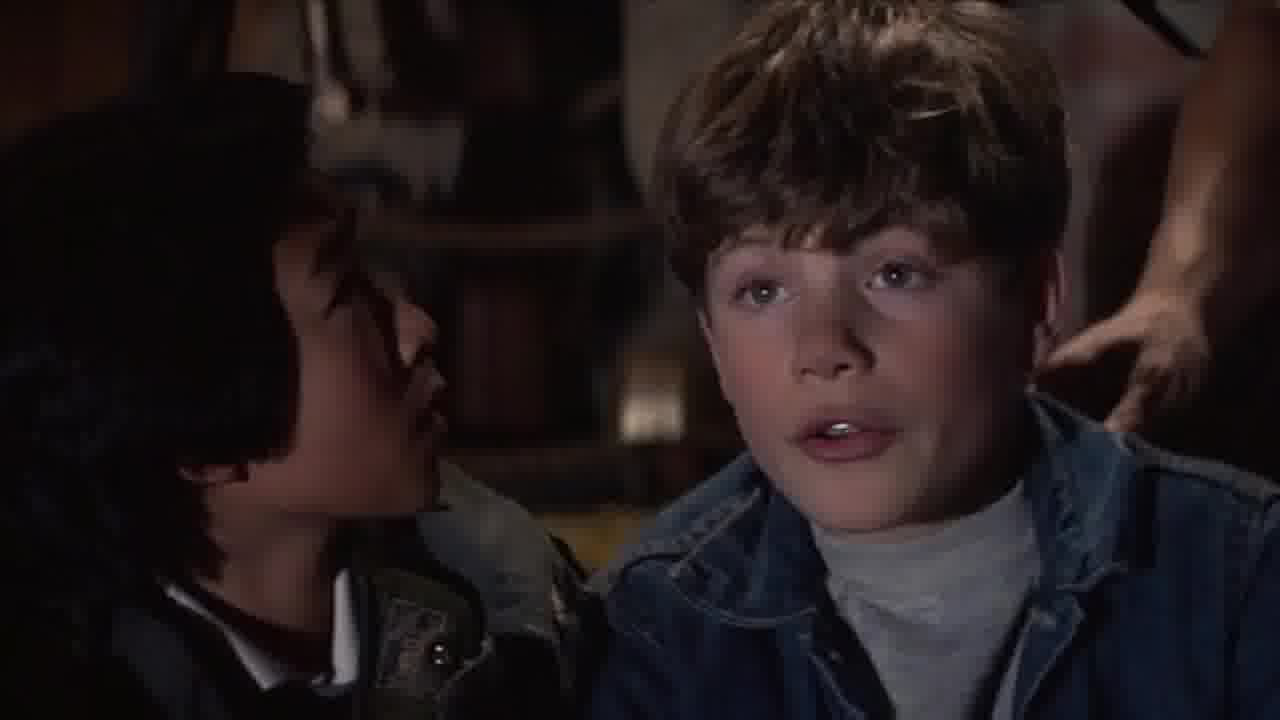}
  \end{subfigure}
  \begin{subfigure}[b]{0.49\textwidth}
  \includegraphics[width=\textwidth]{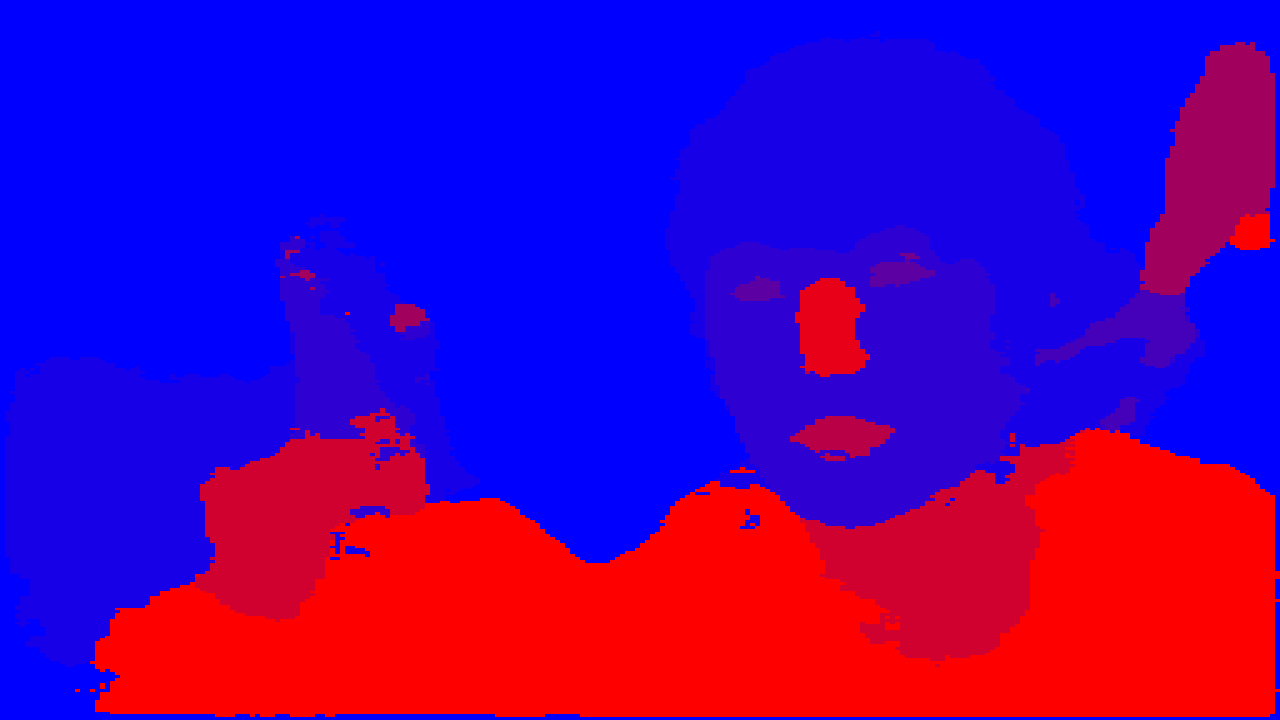}
  \end{subfigure}
    \begin{subfigure}[b]{\textwidth}
  \caption{}
  \end{subfigure}

  \caption{Example visualizations of human body part segmentations for movie frames.}
  \label{fig:examples}
\end{figure}

Example segmentation results for movie frames, not from the Pascal-Parts dataset are shown in Figure \ref{fig:examples}. We tested the trained model on the subset of the Pascal-Parts test set containing "person" objects, and evaluated the results using the Dice score  $\frac{2 TP } { 2TP + FP + FN }$ for each class. In this measure, a true positive (TP) is a correctly labelled pixel of that class, a true negative is (TN) is a correctly labelled pixel not belonging to that class, and false positive (FP) and false negative (FN) are the two possible mislabellings. The Dice scores for each segmentation class are shown in Table \ref{tab:dice}.

\begin{table}[h]
 \caption{Pascal-Parts Test Set Performance}
  \centering
  \begin{tabular}{lll}
    \toprule
    Part Name & Dice \\
    \midrule
    Hair & 0.58  \\
    Head     & 0.60  \\
    Ear     & 0.54 \\
    Eye     & 0.62 \\
    Eyebrow & 0.60 \\ 
    Leg     & 0.38 \\
    Arm     & 0.50 \\
    Mouth     & 0.62  \\
    Neck     & 0.52  \\
    Nose     & 0.57  \\
    Torso & 0.54  \\
    Background & 0.95\\
    \bottomrule
  \end{tabular}
  \label{tab:dice}
\end{table}

\section{Discussion}

The overall performance of the segmentation network is adequate for our purposes, given the sheer number of video frames in a typical movie. Over many frames, the average number fixations to each body part should be a robust estimate. Segmentation quality can be reduced in the presence of occlusion (see Figure \ref{fig:examples}c) and when faces are presented in a frontal view (see Figure \ref{fig:examples}d).  We make this tool available in the hope that it will be helpful to other researchers, and without explicit performance guarantees in any setting.

\bibliographystyle{unsrt}  
\bibliography{bib}

\end{document}